\documentclass[conference]{IEEEtran}
\IEEEoverridecommandlockouts

\usepackage{cite}
\usepackage{amsmath,amssymb,amsfonts}
\usepackage{algorithmic}
\usepackage{graphicx}
\usepackage{textcomp}
\usepackage{xcolor}
\usepackage{tikz}
\usetikzlibrary{shapes.geometric, arrows.meta, positioning, shadows, fit, shapes.symbols}
\usepackage{hyperref}
\usepackage{enumitem}   
\usepackage{listings}   
\hypersetup{
    colorlinks=true,
    linkcolor=blue,
    filecolor=magenta,      
    urlcolor=blue,
    pdftitle={WebNav: An Intelligent Agent for Voice-Controlled Web Navigation},
    pdfpagemode=FullScreen,
    }

\def\BibTeX{{\rm B\kern-.05em{\sc i\kern-.025em b}\kern-.08em
T\kern-.1667em\lower.7ex\hbox{E}\kern-.125emX}}
\begin{document}

\title{WebNav: An Intelligent Agent for Voice-Controlled Web Navigation
}

\author{
\IEEEauthorblockN{Trisanth Srinivasan}
\IEEEauthorblockA{\textit{Dept. of Applied AI} \\
\textit{Cyrion Labs}\\
Dallas, USA \\
trisanth@cyrionlabs.org}
\and
\IEEEauthorblockN{Santosh Patapasti}
\IEEEauthorblockA{\textit{Dept. of Applied AI} \\
\textit{Cyrion Labs}\\
Dallas, USA \\
santosh@cyrionlabs.org}
}

\maketitle

\begin{abstract}
The current state of modern web interfaces, especially in regards to accessibility focused usage is extremely lacking. Traditional methods for web interaction, such as scripting languages and screen readers, often lack the flexibility to handle dynamic content or the intelligence to interpret high-level user goals. To address these limitations, we introduce WebNav, a novel agent for multi-modal web navigation. WebNav leverages a dual Large Language Model (LLM) architecture to translate natural language commands into precise, executable actions on a graphical user interface. The system combines vision-based context from screenshots with a dynamic DOM-labeling browser extension to robustly identify interactive elements. A high-level 'Controller' LLM strategizes the next step toward a user's goal, while a second 'Assistant' LLM generates the exact parameters for execution. This separation of concerns allows for sophisticated task decomposition and action formulation. Our work presents the complete architecture and implementation of WebNav, demonstrating a promising approach to creating more intelligent web automation agents.
\end{abstract}
\begin{IEEEkeywords}
Large Language Models, Web Automation, Human-Computer Interaction, Autonomous Agents, Multi-Modal AI, Assistive Technology
\end{IEEEkeywords}
\section{Introduction}
The World Wide Web has grown into a vital platform for accessing information, conducting commerce, and facilitating social interactions. As web applications have become increasingly sophisticated and dynamic, their inherent complexity can pose significant challenges to users, particularly the millions worldwide with visual impairments \cite{WHO2019Vision}. Navigating intricate menus, completing multi-step forms, and interacting with unconventional UI elements that do not adhere to accessibility guidelines \cite{W3C2018WCAG} often become arduous tasks. Traditional assistive technologies, such as screen readers, typically present information in a linear format, which can be inadequate for executing complex, goal-oriented tasks on modern dynamic web applications \cite{Hailpern2009Web2}. Additionally, conventional automation tools, like Selenium, require programming skills and produce fragile scripts that frequently fail when web page layouts change \cite{Tang2024Steward}.

Consequently, there is a compelling need for a more intelligent paradigm for web interaction—one that enables users to specify high-level objectives in natural language and have an agent autonomously carry out the required actions. The advent of powerful Large Language Models (LLMs) like PaLM \cite{Chowdhery2022PaLM} and GPT-4 \cite{Bubeck2023GPT4} has opened promising avenues to address this need.

In this paper, we introduce WebNav, an innovative multi-modal agent designed to automate web tasks based on natural language. WebNav processes web pages visually, akin to a human user, by interpreting screenshots. To translate visual perceptions into executable actions, WebNav incorporates a custom browser extension that assigns numerical labels to all interactive Document Object Model (DOM) elements, providing a stable, structured reference framework.

A central innovation in WebNav is its two-stage LLM decision-making architecture, which adopts a deliberate "thinking" then "acting" paradigm, conceptually similar to Chain-of-Thought \cite{Wei2022CoT} or Tree-of-Thoughts \cite{Yao2023TreeOfThoughts} reasoning. In the first stage, the \textbf{Controller}, a high-level LLM, analyzes the user's goal, the current visual state, and historical actions to determine the next logical step. Subsequently, the \textbf{Assistant}, a specialized LLM, takes the Controller's instruction and translates it into a precise, structured JSON command suitable for direct execution. This dual-LLM framework effectively decomposes the problem of web navigation into strategic reasoning and detailed action formulation, enhancing robustness and modularity.

The primary contributions of this research include: the design of WebNav as a practical multi-modal agent; a novel two-stage LLM pipeline for task decomposition; and a dynamic DOM-labeling mechanism for grounding actions in the UI context. Finally, we provide a comprehensive, open-source implementation demonstrating the system's effectiveness across diverse web-based tasks. This paper elaborates on WebNav’s architecture and implementation, positioning it as an important advancement toward more capable and autonomous web navigation agents.
\section{Related Work}
Research in automated web interaction has a long history, evolving from programmatic scripting to intelligent, LLM-driven agents.

\subsection{Traditional Web Automation}
Frameworks like Selenium \cite{selenium2024} and Playwright \cite{playwright2024} are the industry standard for web automation. They provide APIs to programmatically interact with DOM elements, but they require significant technical expertise and produce brittle scripts that fail when the UI changes \cite{Tang2024Steward}. This makes them unsuitable for general-purpose, goal-oriented assistance.

\subsection{Assistive Technologies}
Screen readers such as JAWS \cite{jaws2024} and NVDA \cite{nvda2024} are essential tools for users with visual impairments. They serialize the DOM into an audio stream for linear navigation. However, they are not designed for executing complex, multi-step tasks. WebNav's approach is complementary; instead of reading the interface, it aims to operate it on behalf of the user.

\subsection{LLM-Powered Web Agents}
The emergence of powerful LLMs has catalyzed new research into autonomous web agents. Early work demonstrated LLMs generating test scripts from natural language prompts \cite{Ferreira2025LLMTests, Yu2023MobileTests}, a trend surveyed in \cite{Sherifi2025LLMTesting}. More recent systems operate directly on web interfaces. For instance, Mind2Web \cite{Deng2023Mind2Web} provides a comprehensive benchmark for training generalist web agents. Other notable systems include WebGUM \cite{Furuta2024WebGUM}, which focuses on multimodal grounding, and AutoWebGLM \cite{Lai2024AutoWebGLM}, which has shown strong performance on web navigation benchmarks.

Many of these agents, including those based on GPT-4V \cite{OpenAI2023GPT4V}, often employ a single, monolithic LLM in a Reason-Act (ReAct) loop \cite{Yao2023ReAct}. The model is prompted to reason and act in one step, placing a high cognitive load on it. Our WebNav system distinguishes itself with its two-stage Controller-Assistant pipeline, which explicitly separates high-level strategic planning from low-level parameter generation. This modular approach aims to improve reliability. Furthermore, WebNav's use of a dedicated browser extension for explicit element labeling provides a more robust grounding mechanism than relying solely on visual bounding box detection.
\section{System Architecture}
WebNav is a modular system integrating user input, visual perception, and LLM-driven decision-making. The architecture, shown in Fig.~\ref{fig:architecture}, comprises five main components: an Input Handler, Browser Labeling Extension, Screenshot Module, a dual-LLM pipeline, and an Inference Module.

\begin{figure}[t]
\centering
\begin{tikzpicture}[
    font=\small,
    node distance=0.5cm and 0.4cm,
    process/.style={rectangle, rounded corners, draw=black, fill=blue!10, thick, text width=2.2cm, align=center, drop shadow},
    llm/.style={cloud, draw=orange, fill=orange!20, thick, text width=1.8cm, align=center, drop shadow},
    data/.style={trapezium, trapezium left angle=70, trapezium right angle=110, draw=green!60!black, fill=green!10, thick, text width=2cm, align=center},
    browser/.style={rectangle, draw=red!80!black, fill=red!10, thick, text width=2.5cm, align=center, drop shadow},
    pipeline/.style={rectangle, rounded corners, draw=black, dashed, thick, fill=gray!5, inner sep=0.3cm},
    arrow/.style={-Stealth, thick}
]

\node[process] (input) {\textbf{Input Handler}};
\node[process, below=of input] (screenshot) {\textbf{Screenshot Module}};
\node[llm, below=of screenshot, yshift=-0.2cm] (controller) {\textbf{Controller}};
\node[llm, below=of controller, yshift=-0.1cm] (assistant) {\textbf{Assistant}};
\node[process, below=of assistant, yshift=-0.2cm] (inference) {\textbf{Inference Module}};
\node[browser, below=of inference] (browser) {Web Browser};

\node[data, right=of screenshot, xshift=0.5cm] (ss_data) {Labeled \& Unlabeled Screenshots};
\node[process, left=of screenshot, xshift=-0.5cm] (extension) {Browser Extension};

\draw[arrow] (input) -- node[left] {Goal} (screenshot);
\draw[arrow] (screenshot) -- (ss_data);
\draw[arrow] (ss_data) |- (controller);
\draw[arrow] (controller) -- node[right, text width=1cm] {HL Command} (assistant);
\draw[arrow] (ss_data) |- (assistant.east);
\draw[arrow] (assistant) -- node[right] {JSON} (inference);
\draw[arrow] (inference) -- node[left] {Action} (browser);

\draw[arrow, dashed] (screenshot.west) -- (extension.east);
\draw[arrow, dashed] (extension.south) |- (browser.west);
\draw[arrow, dashed, bend right] (browser.north) to (screenshot.south);

\node[pipeline, fit=(controller)(assistant), label={[xshift=-0.1cm, yshift=0.2cm]above:\textbf{LLM Pipeline}}] {};

\end{tikzpicture}
\caption{The condensed WebNav System Architecture. The flow proceeds vertically from user input to browser action, with screenshots feeding the dual-LLM pipeline and a browser extension providing labels.}
\label{fig:architecture}
\end{figure}

\paragraph{User Interaction and Input Handler}
The main loop in \texttt{main.py} handles user input, accepting both typed text and spoken language transcribed via the Whisper model \cite{Radford2022Whisper}. It uses \texttt{pyttsx3} for voice feedback and only acts when invoked by an activation command like “\emph{activate DIGNAV}”.

\paragraph{Vision-based Grounding: The Labeling Extension}
A custom Chrome extension reliably identifies interactive elements. When triggered by \texttt{Alt+Shift+L}, its content script (\texttt{content.js}) overlays the page, querying the DOM for elements like buttons and links. Each visible element is boxed and assigned a unique number, creating a machine-readable map of the UI.

\paragraph{Screenshot Module}
The \texttt{screenshot.py} module captures the visual state in two steps: 1) it saves an “unlabeled” screenshot, and 2) it programmatically triggers the labeling hotkey and captures a second, “labeled” screenshot. This pair provides the LLMs with both visual context and an actionable element map.

\paragraph{The Two-Stage LLM Decision Pipeline}
The system's intelligence resides in a dual-LLM pipeline that processes the goal and screenshots.
The \textbf{Controller Module}, a Gemini-family model in a ReAct-style framework, acts as the high-level strategist. It receives the user goal, screenshots, and action history, then outputs a concise command like \texttt{Click [13]} or \texttt{END}.
The \textbf{Assistant Module}, also a Gemini model, converts the Controller's command into a strict JSON object. For example, \texttt{Click [13]} becomes \texttt{\{"function": "Click", "parameters": \{"number": 13\}\}}. This separation of concerns—"what to do" versus "how to format it"—enhances reliability and minimizes errors.

\paragraph{Inference and Action Execution}
Finally, the \texttt{inference.py} module translates the Assistant’s JSON into actions using \texttt{pyautogui}. The primary functions are \textbf{Click}, which types an element's number to trigger the extension's listener, and \textbf{Type}, which focuses an element before entering text. The module also supports extensible utility functions like \texttt{take\_notes}.
\section{Implementation Details}
The WebNav prototype couples a Python backend with a Chrome extension.

\paragraph{Core System and LLM API}
The Python~3 backend orchestrates the workflow. All model calls are served through Google Gemini endpoints via the \texttt{google.generativeai} SDK. The vision component is conceptually based on a Vision Transformer (ViT) \cite{Dosovitskiy2021ViT} architecture, leveraging principles from contrastive language-image pre-training (CLIP) \cite{Radford2021CLIP}.

\paragraph{GUI Automation and Speech Interface}
Low-level desktop interaction is handled by \texttt{PyAutoGUI} for sending keystrokes and \texttt{mss} for fast screenshot capture. The speech interface uses \texttt{speech\_recognition} with a Whisper backend \cite{Radford2022Whisper} for STT and \texttt{pyttsx3} for offline TTS feedback.

\paragraph{Chrome Extension}
A lightweight extension provides DOM access. The \texttt{manifest.json} registers permissions and a global shortcut, a \texttt{background.js} script relays events, and a \texttt{content.js} script enumerates clickable elements, overlays numeric labels, and triggers clicks as instructed by the agent.
\section{Discussion, Limitations, and Future Work}
While WebNav introduces a robust architecture, its prototype implementation has limitations that guide future work. Its reliance on \texttt{pyautogui} is brittle, requiring window focus and being sensitive to layout changes. The system operates in an open loop, assuming actions succeed without visual verification. Further, it is constrained to the current viewport, lacking a mechanism for scrolling.

The most critical next step is implementing closed-loop verification, where the agent analyzes the outcome of its actions to confirm success and self-correct. To improve reliability, we will replace \texttt{pyautogui} with more direct browser control methods like the Chrome DevTools Protocol (CDP).

Future work also includes enhancing navigational capabilities with intelligent scrolling and incorporating a sophisticated memory module to retain context across multi-step tasks, inspired by approaches that remember and reflect on past actions \cite{Huang2025R2D2}. Rigorous empirical evaluation on standardized benchmarks like Mind2Web \cite{Deng2023Mind2Web} is paramount to measure performance and user satisfaction. Finally, the long-term vision is to implement the proposed \emph{PHYSNAV} capabilities, integrating with real-world sensors to create a unified agent for both digital and physical environments.
\section{Conclusion}
In this paper, we have introduced WebNav, a novel agent for automating web-based tasks through natural language. We have detailed its unique architecture, which centers on a dynamic DOM-labeling browser extension for visual grounding and an innovative two-stage LLM pipeline for decision-making. The Controller LLM acts as a high-level strategist, while the Assistant LLM ensures precise, machine-readable action formatting. This separation of concerns represents a promising method for improving the reliability and capability of LLM-powered agents.

While still in a theoretical and developmental stage, the complete implementation of WebNav demonstrates the feasibility and potential of this approach. It lays the groundwork for creating more intelligent, robust, and accessible systems that can bridge the gap between human intent and complex digital interfaces. Future work will focus on implementing closed-loop verification and conducting rigorous user studies to empirically validate the benefits of our architecture. Ultimately, WebNav is a step toward a future where anyone can command a digital agent to perform complex tasks as easily as having a conversation.

\bibliographystyle{IEEEtran}
\bibliography{references}

\begin{thebibliography}{10}
\providecommand{\url}[1]{#1}
\csname url@samestyle\endcsname
\providecommand{\newblock}{\relax}
\providecommand{\bibinfo}[2]{#2}
\providecommand{\BIBentrySTDinterwordspacing}{\spaceskip=0pt\relax}
\providecommand{\BIBentryALTinterwordstretchfactor}{4}
\providecommand{\BIBentryALTinterwordspacing}{\spaceskip=\fontdimen2\font plus
\BIBentryALTinterwordstretchfactor\fontdimen3\font minus \fontdimen4\font\relax}
\providecommand{\BIBforeignlanguage}[2]{{%
\expandafter\ifx\csname l@#1\endcsname\relax
\typeout{** WARNING: IEEEtran.bst: No hyphenation pattern has been}%
\typeout{** loaded for the language `#1'. Using the pattern for}%
\typeout{** the default language instead.}%
\else
\language=\csname l@#1\endcsname
\fi
#2}}
\providecommand{\BIBdecl}{\relax}
\BIBdecl

\bibitem{WHO2019Vision}
W.~H. Organization, \emph{World Report on Vision}.\hskip 1em plus 0.5em minus 0.4em\relax World Health Organization, 2019.

\bibitem{W3C2018WCAG}
{W3C Web Accessibility Initiative}, ``Web content accessibility guidelines (wcag) 2.1,'' 2018, w3C Recommendation; \url{https://www.w3.org/TR/WCAG21/}.

\bibitem{Hailpern2009Web2}
J.~Hailpern, R.~Ladner, A.~Lisin \emph{et~al.}, ``Web 2.0: Blind to an accessible new world,'' in \emph{Proc. 18th Int’l ACM SIGACCESS Conf. on Computers and Accessibility (W4A)}, 2009, pp. 1--8.

\bibitem{Tang2024Steward}
B.~Tang and H.~Shin, ``Steward: Natural language web automation,'' \emph{arXiv preprint arXiv:2409.15441}, 2024.

\bibitem{Chowdhery2022PaLM}
A.~Chowdhery, S.~Narang, J.~Devlin, M.~Bosma, G.~Mishra, A.~Roberts, P.~Barham, H.-W. Chung, C.~Sutton \emph{et~al.}, ``Palm: Scaling language modeling with pathways,'' \emph{arXiv preprint arXiv:2204.02311}, 2022.

\bibitem{Bubeck2023GPT4}
S.~Bubeck, V.~Chandrasekaran, R.~Eldan, J.~Gehrke, E.~Horvitz, E.~Kamar, P.~Lee, Y.-T. Lee, Y.~Li, S.~Lundberg \emph{et~al.}, ``Sparks of artificial general intelligence: Early experiments with gpt-4,'' \emph{arXiv preprint arXiv:2303.12712}, 2023.

\bibitem{Wei2022CoT}
J.~Wei, X.~Wang, D.~Schuurmans, M.~Bosma, B.~Chi, S.~Le, H.~Zhou, D.~Song, J.~Feng, D.~Zhang \emph{et~al.}, ``Chain-of-thought prompting elicits reasoning in large language models,'' in \emph{Advances in Neural Information Processing Systems 35}, 2022, pp. 24\,824--24\,837.

\bibitem{Yao2023TreeOfThoughts}
S.~Yao, R.~Ye, R.~Chintala, C.~Dathathri, J.~Sharma, A.~Jain, and P.~Liang, ``Tree of thoughts: Deliberate problem solving with large language models,'' in \emph{Advances in Neural Information Processing Systems (NeurIPS) 36}, 2023, pp. 4110--4126.

\bibitem{selenium2024}
``Selenium webdriver documentation,'' \url{https://www.selenium.dev/documentation/webdriver/}, 2024, [Online; accessed 21-May-2024].

\bibitem{playwright2024}
``Playwright documentation,'' \url{https://playwright.dev/docs/intro}, 2024, [Online; accessed 21-May-2024].

\bibitem{jaws2024}
{Freedom Scientific}, ``Jaws (job access with speech),'' \url{https://www.freedomscientific.com/Products/Blindness/JAWS}, 2024, [Online; accessed 21-May-2024].

\bibitem{nvda2024}
{NV Access}, ``Nonvisual desktop access (nvda),'' \url{https://www.nvaccess.org/}, 2024, [Online; accessed 21-May-2024].

\bibitem{Ferreira2025LLMTests}
M.~Ferreira, L.~Viegas, J.~P. Faria, and B.~Lima, ``Acceptance test generation with large language models: An industrial case study,'' \emph{arXiv preprint arXiv:2504.07244}, 2025, presented at SPSNA 2024 workshop; supplementary materials available.

\bibitem{Yu2023MobileTests}
Z.~Yu, J.~Wei, K.~Wang, and et~al., ``Application of large language models in mobile app test generation,'' \emph{arXiv preprint arXiv:2307.04724}, 2023.

\bibitem{Sherifi2025LLMTesting}
A.~Sherifi, C.~Pan, S.~Ding, S.~Pan, and D.~Lo, ``A survey on the application of large language models in software testing,'' \emph{arXiv preprint arXiv:2501.00217}, 2025.

\bibitem{Deng2023Mind2Web}
X.~Deng, Y.-C. Hsiao, H.~Behl, S.~Ghosh, S.~Iyer, P.-C. Hsieh, R.~Prasaath, S.~Gummadi, C.~Li, M.~Sundaram, J.~Gao, K.~Narasimhan, R.~Agarwal, H.~Firooz, G.~Bansal, T.-K.~S. Kumar, and D.~Lange, ``Mind2web: Towards a generalist agent for the web,'' in \emph{Advances in Neural Information Processing Systems (NeurIPS) 36}, 2023, pp. 22\,179--22\,191.

\bibitem{Furuta2024WebGUM}
H.~Furuta, K.-H. Lee, O.~Nachum, J.~Hwang, S.~Jin, L.~Song, and D.-Y. Kim, ``Webgum: Multimodal web navigation with instruction-finetuned foundation models,'' \emph{arXiv preprint arXiv:2305.11854}, 2023, accepted at ICLR 2024.

\bibitem{Lai2024AutoWebGLM}
H.~Lai, X.~Liu, I.~Iong, S.~Yao, Y.~Chen, P.~Shen, H.~Yu, H.~Zhang, X.~Zhang, Y.~Dong, and J.~Tang, ``Autowebglm: A large language model-based web navigating agent,'' in \emph{Proc. 30th ACM SIGKDD Int’l Conf. on Knowledge Discovery and Data Mining (KDD)}, 2024, pp. ---.

\bibitem{OpenAI2023GPT4V}
{OpenAI}, ``Gpt-4v(ision) system card,'' Tech. Rep., 2023, arXiv preprint arXiv:2309.17421.

\bibitem{Yao2023ReAct}
S.~Yao, Y.~Zhao, D.~Bommasani, and P.~Liang, ``React: Synergizing reasoning and acting in language models,'' in \emph{Advances in Neural Information Processing Systems (NeurIPS) 36}, 2023, pp. 26\,639--26\,653.

\bibitem{Radford2022Whisper}
A.~Radford, J.~Wu, R.~Child, D.~Luan, T.~Amodei, and I.~Sutskever, ``Robust speech recognition via large-scale weak supervision,'' \emph{arXiv preprint arXiv:2212.04356}, 2022.

\bibitem{Dosovitskiy2021ViT}
A.~Dosovitskiy, L.~Beyer, A.~Kolesnikov, D.~Weissenborn, X.~Zhai, T.~Unterthiner, M.~Dehghani, N.~Minderer, G.~Heigold, S.~Gelly, J.~Houlsby, and O.~Vinyals, ``An image is worth 16x16 words: Transformers for image recognition at scale,'' in \emph{Proc. Int. Conf. on Learning Representations (ICLR)}, 2021.

\bibitem{Radford2021CLIP}
A.~Radford, J.~W. Kim, C.~Hallacy, A.~Ramesh, G.~Goh, S.~Agarwal, G.~Sastry, A.~Askell, P.~Mishkin, J.~Clark \emph{et~al.}, ``Learning transferable visual models from natural language supervision,'' in \emph{Proc. Int. Conf. on Machine Learning (ICML)}, 2021.

\bibitem{Huang2025R2D2}
T.~Huang, M.~Basu, T.~Ward, and D.~Khashabi, ``R2d2: Remembering, reflecting, and dynamic decision making for web agents,'' \emph{arXiv preprint arXiv:2501.12485}, 2025.

\end{thebibliography}

\end{document}